\documentclass[11pt]{article}
\usepackage{acl2016}
\pdfoutput=1
\usepackage{times}
\usepackage{url}
\usepackage{latexsym}
\usepackage{amsmath}
\usepackage{graphicx}
\usepackage{color}
\usepackage{multirow}
\usepackage[utf8]{inputenc}
\usepackage{algorithm}
\usepackage{algorithmicx}
\usepackage{algpseudocode}
\usepackage{pifont}

\newcommand{\cmark}{\ding{51}}%
\newcommand{\xmark}{\ding{55}}%

\def\Sref#1{Section~\ref{#1}} 
\def\Tref#1{Table~\ref{#1}}
\def\Fref#1{Figure~\ref{#1}}

\def\footurl#1{\footnote{\url{#1}}}

\aclfinaltrue

\DeclareMathOperator*{\fv}{fv}
\DeclareMathOperator*{\GEN}{GEN}

\makeatletter
\newcommand{\@BIBLABEL}{\@emptybiblabel}
\newcommand{\@emptybiblabel}[1]{}
\makeatother
\usepackage{hyperref}

\begin{document}

\title{Target-Side Context for Discriminative Models in Statistical Machine Translation}

\author{Aleš Tamchyna$^{1,2}$, Alexander Fraser$^1$, Ondřej Bojar$^2$ and Marcin Junczys-Dowmunt$^3$\\
  {\normalsize$^1$LMU Munich, Munich, Germany}\\
  {\normalsize$^2$Charles University in Prague, Prague, Czech Republic}\\
  {\normalsize$^3$Adam Mickiewicz University in Poznań, Poznań, Poland}\\
{\normalsize \{tamchyna,bojar\}@ufal.mff.cuni.cz~~~~fraser@cis.uni-muenchen.de~~~~junczys@amu.edu.pl}\\
}

\maketitle

\begin{abstract}
Discriminative 
translation
models 
utilizing
source context have been 
shown to help
statistical machine
translation
performance.
We propose a novel extension of this work using
target context information. Surprisingly, we show that
this model can be efficiently integrated
directly in the decoding process. 
Our
approach scales to large training data sizes and 
results in consistent
improvements 
in
translation quality
on four language pairs.
We also provide an analysis comparing the strengths of the baseline
source-context model with our extended source-context and target-context
model and we show that our
extension 
allows us 
to better capture
morphological coherence. Our work 
is
freely available as part of Moses.
\end{abstract}

%
%

\section{Introduction}

Discriminative lexicons 
address some of the core challenges of phrase-based
MT (PBMT)
when translating to morphologically rich languages, such as Czech, namely sense disambiguation and morphological coherence.
The first issue is semantic: given a source word or phrase, which of its
possible meanings (i.e., which stem or lemma) should we choose? Previous work has shown that this can be addressed using a discriminative lexicon.
The second issue has to do with morphology
(and syntax): given that we selected the correct meaning, which of its 
inflected 
surface
forms is appropriate? 
In this work, we integrate such a model directly into the
SMT decoder. This enables our classifier to extract features not only from the
full source sentence but also from a limited target-side context. This allows
the model to not only help with semantics but also to improve morphological and
syntactic coherence.


For \emph{sense disambiguation}, source context is the main source of
information, as has been shown in previous work \cite{Vickrey-wsd}, \cite{carpuat:07b},
\cite{gimpel-08} inter alia.
Consider the first set of examples in \Fref{pbmt-outputs}, produced by a
strong baseline PBMT system. The English word ``shooting'' has multiple senses
when translated into Czech: it may either be the act of firing a weapon or
making a film. When the cue word ``film'' is close, the phrase-based model is
able to use it in one phrase with the ambiguous ``shooting'', disambiguating
correctly the translation. When we add a single
word in between, the model fails to capture the relationship and the most
frequent sense is selected instead. Wider source context information is required
for correct disambiguation.

\begin{figure*}[t]
\begin{center}
\begin{tabular}{lll}
  \bf Input & \bf PBMT Output & \\
  \hline
  shooting of the film . & natáčení filmu . & \cmark \\
                         & \emph{\textbf{shooting$_{camera}$} of\_film .} \\
  shooting of the expensive film . & střelby na drahý film . & \xmark \\
                                   & \emph{\textbf{shootings$_{gun}$} at expensive film .} \\
	\hline
	the man saw a cat . & muž uviděl kočku . & \cmark \\
											&  \emph{man saw \textbf{cat$_{acc}$}} . \\
	the man saw a black cat . & muž spatřil černou kočku . & \cmark \\
											&  \emph{man saw \textbf{black$_{acc}$ cat$_{acc}$}} . \\
	the man saw a yellowish cat . & muž spatřil nažloutlá kočka . & \xmark \\
											&  \emph{man saw \textbf{yellowish$_{nom}$ cat$_{nom}$}} . \\
\end{tabular}
\end{center}
\caption{Examples of problems of PBMT: lexical selection and morphological coherence. Each translation has a corresponding gloss in italics.}
\label{pbmt-outputs}
\end{figure*}

While word/phrase senses can usually be inferred from the source sentence, the
correct \emph{selection of surface forms} requires also information from the
target.  Note that we can obtain \emph{some} information from the source. For
example, an English subject is often translated into a Czech subject; in which
case the Czech word should be in nominative case. But there are many decisions
that happen during decoding which determine morphological and syntactic
properties of words -- verbs can have translations which differ in valency
frames, they may be translated in either active or passive voice (in which case
subject and object would be switched), nouns may have different possible
translations which differ in gender, etc.

The correct selection of surface forms plays a crucial role in preserving
meaning in morphologically rich languages because it is morphology rather than word
order that expresses relations between words. (Word order tends to be relatively free
and driven more by semantic constraints rather than syntactic constraints.)

The language model is only partially able to capture this phenomenon. It has a
limited scope and perhaps more seriously, it suffers from data sparsity.  The
units captured by both the phrase table and the LM are mere sequences of words.
In order to estimate their probability, we need to observe them in the training
data (many times, if the estimates should be reliable). However, the number of
possible $n$-grams grows exponentially as we increase $n$, leading to
unrealistic requirements on training data sizes. This implies that the current models can
(and often do) miss relationships between words even within their theoretical
scope.

The second set of sentences in \Fref{pbmt-outputs} demonstrates the problem of
data sparsity for morphological coherence. While the phrase-based system can
correctly transfer the morphological case of ``cat'' and even ``black cat'', the
less usual ``yellowish cat'' is mistranslated into nominative case, even though
the correct phrase ``yellowish $|||$ nažloutlou'' exists in the phrase table.  A
model with a suitable representation of two preceding words could easily infer
the correct case in this example. 



Our contributions are the following:

\begin{itemize}

  \item We show that the addition of a feature-rich discriminative model significantly
    improves translation quality even for large data sizes and that target-side
    context information consistently further increases this improvement.
    
  \item We provide an analysis of the outputs which confirms that source-context
    features indeed help with semantic disambiguation (as is well known).
    Importantly, we also show that our novel use of
    target context improves morphological and syntactic coherence.

  \item In addition to extensive experimentation on translation from
    English to Czech, we also evaluate English to German, English to
    Polish and English to Romanian tasks, with improvements on
    translation quality in all tasks, showing that our work is broadly
    applicable.

  \item We describe several optimizations which allow target-side
    features to be used efficiently in the context of phrase-based
    decoding.

  \item 
    Our implementation is freely available in the widely used
    open-source MT toolkit Moses, enabling other researchers to
    explore discriminative modelling with target context in MT.

\end{itemize}

%
%

\section{Discriminative Model with Target-Side Context}

Several different ways of using feature-rich models in MT have been proposed,
see \Sref{related-work}. We describe our approach in this section.

\subsection{Model Definition}

Let $f$ be the source sentence and $e$ its translation. We denote
source-side phrases (given a particular phrasal segmentation) $(\bar{f_1},
\ldots, \bar{f_m})$ and the individual words $(f_1, \ldots, f_n)$. We use a
similar notation for target-side words/phrases.

For simplicity, let $e_{prev}, e_{prev-1}$ denote the words preceding the
current target phrase. Assuming target context size of two, we model the
following probability distribution:

\begin{equation}
P(e|f) \propto \prod_{(\bar{e_i}, \bar{f_i}) \in (e,f)}
P(\bar{e_i}|\bar{f_i}, f, e_{prev}, e_{prev-1})
\end{equation}

The probability of a translation is the product of phrasal translation
probabilities which are conditioned on the source phrase, the full source
sentence and several previous target words.

Let $\GEN(\bar{f_i})$ be the set of possible translations of the source phrase
$\bar{f_i}$ according to the phrase table. We also define a ``feature vector''
function $\fv(\bar{e_i}, \bar{f_i}, f, e_{prev}, e_{prev-1})$ which outputs a
vector of features given the phrase pair and its context information. We also
have a vector of feature weights $w$ estimated from the training data. Then our
model defines the phrasal translation probability simply as follows:

\begin{equation}
\begin{split}
&P(\bar{e_i}|\bar{f_i}, f, e_{prev}, e_{prev-1}) \\
&= \frac{\exp(w \cdot \fv(\bar{e_i}, \bar{f_i}, f, e_{prev},e_{prev-1}))}
  {\sum\limits_{\bar{e'}\in \GEN(\bar{f_i})} 
    \exp(w \cdot \fv(\bar{e'}, \bar{f_i}, f, e_{prev}, e_{prev-1}))}
\end{split}
\end{equation}

This definition implies that we have to locally normalize the classifier outputs
so that they sum to one.

In PBMT, translations are usually scored by a log-linear model. Our classifier
produces a single score (the conditional phrasal probability) which we add to
the standard log-linear model as an additional feature. The MT system therefore
does not have direct access to the \emph{classifier} features, only to the final
score.

\begin{table*}[t]
  \begin{center}
    \begin{tabular}{r|lll}
      \multirow{2}{*}{Feature Type} & \multicolumn{3}{c}{Configurations} \\
                                    & Czech & German & Polish, Romanian \\
      \hline
      Source Indicator  & f, l, l+t, t                 & f, l, l+t, t                 & l, t                 \\
      Source Internal   & f, f+a, f+p, l, l+t, t, a+p  & f, f+a, f+p, l, l+t, t, a+p  & l, l+a, l+p, t, a+p  \\
      Source Context    & f (-3,3), l (-3,3), t (-5,5) & f (-3,3), l (-3,3), t (-5,5) & l (-3,3), t (-5,5) \\
      \hline
      Target Context    & f (2), l (2), t (2), l+t (2) & f (2), l (2), t (2), l+t (2) & l (2), t (2) \\
      Bilingual Context & ---                          & l+t/l+t (2)                  & l+t/l+t (2) \\
      \hline
      Target Indicator  & f, l, t                      & f, l, t                      & l, t \\
      Target Internal   & f, l, l+t, t                 & f, l, l+t, t                 & l, t \\
    \end{tabular}
    \caption{List of used feature templates. Letter abbreviations refer to word
    factors: f (form), l (lemma), t (morphological tag), a (analytical
  function), p (lemma of dependency parent). Numbers in parentheses indicate
context size.}
    \label{feature-set}
  \end{center}
\end{table*}

\subsection{Global Model}

We use 
the
Vowpal Wabbit (VW) classifier\footurl{http://hunch.net/~vw/} in this
work. 
\newcite{tamchyna14:pbml} already integrated VW into Moses. We started
from their implementation in order to carry out our work.
Classifier features 
are divided
into two ``namespaces'': 
\begin{itemize}
  \item {\bf S.} Features that do not depend on the current phrasal translation
    (i.e., source- and target-context features).
  \item {\bf T.} Features of the current phrasal translation.
\end{itemize}

We make heavy use of feature processing available in VW, namely quadratic
feature expansions and label-dependent features. When generating features for a
particular set of translations, we first create the \emph{shared} features (in
the namespace $S$). 
These only depend on (source and target) context and are
therefore constant for all possible translations of a given phrase. (Note that
target-side context naturally depends on the current partial translation. However, when we
process the possible translations for a single source phrase, the target context is
constant.)

Then for each translation, we extract its features and store them in the
namespace $T$. Note that we do not provide a label (or class) to VW -- it is up
to these \emph{translation} features to describe the target phrase. (And this is
what is referred to as ``label-dependent features'' in VW.)

Finally, we add the Cartesian product between the two namespaces to the feature
set: every \emph{shared} feature is combined with every \emph{translation}
feature.

This setting allows us to train only a single, global model with powerful
feature sharing. For example, thanks to the label-dependent format, we can
decompose both the source phrase and the target phrase into words and have
features such as {\tt s\_cat\_t\_kočka} which capture phrase-internal word
translations. Predictions for rare phrase pairs are then more robust thanks to
the rich statistics collected for these word-level feature pairs.

\subsection{Extraction of Training Examples}

Discriminative models in MT are typically trained by creating one training
instance per extracted phrase from the entire training data. The target side of
the extracted phrase is a positive label, and all other phrases observed aligned
to the extracted phrase (anywhere in the training data) are the negative labels.

We train our model in a similar fashion: for each sentence in the parallel
training data, we look at all possible phrasal segmentations. Then for each
source span, we create a training example. We obtain the set of possible
translations $\GEN(\bar{f})$ from the phrase table. Because we do not have
actual classes, each translation is defined by its label-dependent features and
we associate a loss with it: 0 loss for the correct translation and 1 for all
others.

Because we train both our model and the standard phrase table on the same
dataset, we use leaving-one-out in the classifier training to avoid
over-fitting. We look at phrase counts and co-occurrence counts in the training
data, we subtract one from the number of occurrences for the current source
phrase, target phrase and the phrase pair.  If the count goes to zero, we skip
the training example.  Without this technique, the classifier might learn to
simply trust very long phrase pairs which were extracted from the same training
sentence. 

For target-side context features, we simply use the true (gold) target
context. This leads to training which is similar to language model
estimation; this model 
is somewhat similar to the
neural joint
model for MT \cite{devlin},
but in our case
implemented using a linear
(maximum-entropy-like) model.

\subsection{Training}

We use Vowpal Wabbit in the {\tt --csoaa\_ldf mc} setting which reduces our
multi-class problem to one-against-all binary classification. We use the
logistic loss as our objective. We experimented with various settings of
L2 regularization but were not able to get an improvement over not using
regularization at all. We train each model with 10 iterations over the data.

We evaluate all of our models on a held-out set. We use the same dataset as for
MT system tuning because it closely matches the domain of our test set. We
evaluate model accuracy after each pass over the training data to detect
over-fitting and we select the model with the highest held-out accuracy.

\subsection{Feature Set}

Our feature set requires some linguistic processing of the data. We use the
factored MT setting \cite{factoredmt} and we represent each type of information
as an individual factor. On the source side, we use the word surface form, its
lemma, morphological tag, analytical function (such as \emph{Subj} for subjects)
and the lemma of the parent node in the dependency parse tree. On the target
side, we only use word lemmas and morphological tags. 



\Tref{feature-set} lists our feature sets for each language pair. We implemented
\emph{indicator} features for both the source and target side; these are simply
concatenations of the words in the current phrase into a single feature.
\emph{Internal} features describe words within the current phrase.
\emph{Context} features are extracted either from a window of a fixed size
around the current phrase (on the source side) or from a limited left-hand side
context (on the target side). \emph{Bilingual context} features are
concatenations of target-side context words and their source-side counterparts
(according to word alignment); these features are similar to bilingual tokens in
bilingual LMs \cite{bilingual-lm}.  Each of our feature types can be configured
to look at any individual factors or their combinations.

The features in \Tref{feature-set} are divided into three sets. The first set
contains label-independent (=shared) features which only depend on the source
sentence.  The second set contains shared features which depend on target-side
context; these can only be used when VW is applied during decoding.  We use
target context size two in all our experiments.\footnote{In preliminary
experiments we found that using a single word was less effective and larger
context did not bring improvements, possibly because of over-fitting.} Finally,
the third set contains label-dependent features which describe the currently
predicted phrasal translation.

Going back to the examples from \Fref{pbmt-outputs}, our model can disambiguate
the translation of ``shooting'' based on the source-context features (either the
full form or lemma). For the morphological disambiguation of the translation of
``yellowish cat'', the model has access to the morphological tags of the
preceding target words which can disambiguate the correct morphological case.

We used slightly different subsets of the full feature set for different
languages. In particular, we left out surface form features and/or bilingual
features in some settings because they decreased performance, presumably due to
over-fitting.

%
%

\section{Efficient Implementation}

Originally, we assumed that using target-side context features in decoding would
be too expensive, considering that we would have to query our model roughly as
often as the language model. In preliminary experiments, we therefore focused on
$n$-best list re-ranking. We obtained small gains but all of our results were
substantially worse than with the integrated model, so we omit them from the
paper.

We find that decoding with a feature-rich target-context model is in fact
feasible. In this section, we describe optimizations at different stages of our
pipeline which make training and inference with our model practical.

\subsection{Feature Extraction}

We implemented the code for feature extraction only once; identical code is used
at training time and in decoding. At training time, the generated features are
written into a file whereas at test time, they are fed directly into the
classifier via its library interface.

This design decision not only ensures consistency in feature representation
but also makes the process of feature extraction efficient. In training, we are
easily able to use multi-threading (already implemented in Moses) and because
the processing of training data is a trivially parallel task, we can also use
distributed computation and run separate instances of (multi-threaded) Moses on
several machines. This enables us to easily produce training files from millions
of parallel sentences within a short time.

\subsection{Model Training}

VW is a very fast classifier by itself, however for very large data, its
training can be further sped up by using parallelization. We take advantage of
its implementation of the \emph{AllReduce} scheme which we utilize in a grid
engine environment. We shuffle and shard the data and then assign each shard to
a worker job. With AllReduce, there is a master job which synchronizes the
learned weight vector with all workers. We have compared this approach with the
standard single-threaded, single-process training and found that we obtain
identical model accuracy. We usually use around 10-20 training jobs. 

This way, we can process our large training files quickly and train the full
model (using multiple passes over the data) within hours; effectively, neither
feature extraction nor model training become a significant bottleneck in the
full MT system training pipeline.

\subsection{Decoding}

In phrase-based decoding, translation is generated from left to right. At each
step, a partial translation (initially empty) is extended by translating a
previously uncovered part of the source sentence. There are typically many ways
to translate each source span, which we refer to as translation options. The
decoding process gradually extends the generated partial translations until the
whole source sentence is covered; the final translation is then the full
translation hypothesis with the highest model score. Various pruning strategies
are applied to make decoding tractable.

Evaluating a feature-rich classifier during decoding is a computationally
expensive operation. Because the features in our model depend on target-side
context, the feature function which computes the classifier score cannot
evaluate the translation options in isolation (independently of the partial
translation). Instead, similarly to a language model, it needs to look at
previously generated words. This also entails maintaining a state which captures
the required context information.

A naive integration of the classifier would simply generate all source-context
features, all target-context features and all features describing the
translation option each time a partial hypothesis is evaluated. This is
a computationally very expensive approach.

We instead propose several technical solutions which make decoding reasonably
fast. Decoding a single sentence with the naive approach takes 13.7 seconds on
average. With our optimization, this average time is reduced to 2.9 seconds,
i.e. almost by 80 per cent. The baseline system produces a translation in 0.8
seconds on average.

{\bf Separation of source-context and target-context evaluation.} Because we
have a linear model, the final score is simply the dot product between a weight
vector and a (sparse) feature vector. It is therefore trivial to separate it
into two components: one that only contains features which depend on the source
context and the other with target context features. We can pre-compute the
source-context part of the score before decoding (once we have all translation
options for the given sentence). We cache these partial scores and when the
translation option is evaluated, we add the partial score of the target-context
features to arrive at the final classifier score.

{\bf Caching of feature hashes.} VW uses feature hashing internally and it is
possible to obtain the hash of any feature that we use. When we encounter a
previously unseen target context (=state) during decoding, we store the hashes
of extracted features in a cache. Therefore for each context, we only run the
expensive feature extraction once. Similarly, we pre-compute feature hash
vectors for all translation options.

{\bf Caching of final results.} Our classifier locally normalizes the scores so
that the probabilities of translations for a given span sum to one. This cannot
be done without evaluating all translation options for the span at the same
time. Therefore, when we get a translation option to be scored, we fetch all
translation options for the given source span and evaluate all of them. We then
normalize the scores and add them to a cache of final results. When the other
translation options come up, their scores are simply fetched from the cache.
This can also further save computation when we get into a previously seen state
(from the point of view of our classifier) and we evaluate the same set of
translation options in that state; we will simply find the result in cache in
such cases.

When we combine all of these optimizations, we arrive at the query
algorithm shown in \Fref{vwdecoding}.

\begin{figure}[h]
  \begin{center}
    \begin{algorithmic}
      \Function{evaluate}{$t, s$}
        \State span = $t$.getSourceSpan()
        \If{\textbf{not} resultCache.has(span, $s$)}
          \State scores = ()
          \If{\textbf{not} stateCache.has($s$)}
            \State stateCache[$s$] = CtxFeatures($s$)
          \EndIf
          \ForAll{$t' \gets$ span.tOpts()}
            \State srcScore = srcScoreCache[$t'$]
            \State $c$.addFeatures(stateCache[$s$])
            \State $c$.addFeatures(translationCache[$t'$])
            \State tgtScore = $c$.predict()
            \State scores[$t'$] = srcScore + tgtScore
          \EndFor
          \State normalize(scores)
          \State resultCache[span, $s$] = scores
        \EndIf
        \State \Return resultCache[span, $s$][$t$]
      \EndFunction
    \end{algorithmic}
    \caption{Algorithm for obtaining classifier predictions during decoding. The
    variable $t$ stands for the current translation, $s$ is the current state
  and $c$ is an instance of the classifier.}
    \label{vwdecoding}
  \end{center}
\end{figure}

%
%

\section{Experimental Evaluation}

We run the main set of experiments on English to Czech translation. To verify
that our method is applicable to other language pairs, we also present
experiments in English to German, Polish, and Romanian.

In all experiments, we use Treex \cite{tectomt:popel:2010} to lemmatize and tag
the source data and also to obtain dependency parses of all English sentences.

\subsection{English-Czech Translation}

As parallel training data, we use (subsets of) the CzEng 1.0 corpus
\cite{czeng10:lrec2012}. For tuning, we use the WMT13 test set \cite{wmt13} and
we evaluate the systems on the WMT14 test set \cite{wmt14}. We lemmatize and tag
the Czech data using Morphodita \cite{morphodita}.

Our baseline system is a standard phrase-based Moses setup. The phrase table in
both cases is factored and outputs also lemmas and morphological tags. We train
a 5-gram LM on the target side of parallel data.

We evaluate three settings in our experiments:

\begin{itemize}
  \item baseline -- vanilla phrase-based system,
  \item +source -- our classifier with source-context features only,
  \item +target -- our classifier with both source-context and target-context features.
\end{itemize}        

For each of these settings, we vary the size of the training data for our
classifier, the phrase table and the LM. We experiment with three different
sizes: small (200 thousand sentence pairs), medium (5 million sentence pairs), 
and full (the whole CzEng corpus, over 14.8 million sentence pairs).

For each setting, we run system weight optimization (tuning) using 
minimum
error
rate training \cite{mert} five times and report the average BLEU score. We use
MultEval \cite{multeval} to compare the systems and to determine whether the
differences in results are statistically significant. We always compare the
baseline with +source and +source with +target.

\Tref{vwscores} shows the obtained results. Statistically significant
differences ($\alpha$=0.01) are marked in bold. The source-context model
does not help in the small data setting but brings a substantial improvement of
0.7-0.8 BLEU points for the medium and full data settings, which is an
encouraging result.

Target-side context information allows our model to push the translation quality
further: even for the small data setting, it brings a substantial improvement of
0.5 BLEU points and the gain remains significant as the data size increases. Even in the full
data setting, target-side features improve the score by roughly 0.2 BLEU points.

Our results demonstrate that feature-rich models scale to large data size both
in terms of technical feasibility and of translation quality improvements.
Target side information seems consistently beneficial, adding further 0.2-0.5 BLEU
points on top of the source-context model.

\begin{table}[h]
  \begin{center}
    \begin{tabular}{l|ccc}
      data size & small & medium & full \\
      \hline 
      baseline & 10.7 & 15.2 & 16.7 \\
      +source & 10.7 & \bf 16.0 & \bf 17.3 \\
      +target & \bf 11.2 & \bf 16.4 & \bf 17.5 \\
    \end{tabular}
    \caption{BLEU scores obtained on the WMT14 test set. We report the
      performance of the baseline, the source-context model and the full model.}
    \label{vwscores}
  \end{center}
\end{table}


\textbf{Intrinsic Evaluation.} For completeness, we report intrinsic evaluation
results. We evaluate the classifier on a held-out set (WMT13 test set) by
extracting all phrase pairs from the test input aligned with the test reference
(similarly as we would in training) and scoring each phrase pair (along with
other possible translations of the source phrase) with our classifier. An
instance is classified correctly if the true translation obtains the highest
score by our model. A baseline which always chooses the most frequent phrasal
translation obtains accuracy of 51.5. For the source-context model, the
held-out accuracy was 66.3, while the target context model achieved accuracy of
74.8. Note that this high difference is somewhat misleading because in this
setting, the target-context model has access to the true target context (i.e.,
it is cheating).

\begin{figure*}[!t]
  \begin{center}
    \begin{tabular}{rl}
      \bf input: & the most intensive mining took place there from 1953 to 1962 . \\
      \bf baseline: & nejvíce intenzivní těžba {\bf došlo} tam z roku 1953 {\bf , aby} 1962 . \\
                    & \emph{the\_most intensive mining$_{nom}$ \textbf{there\_occurred} there from 1953 \textbf{, in\_order\_to} 1962 .} \\
      \bf +source: & nejvíce intenzivní {\bf těžby místo} tam z roku 1953 do roku 1962 . \\
                   & \emph{the\_most intensive \textbf{mining$_{gen}$} \textbf{place} there from year 1953 until year 1962 .} \\
      \bf +target: & nejvíce intenzivní těžba probíhala od roku 1953 do roku 1962 . \\
                   & \emph{the\_most intensive mining$_{nom}$ occurred from year 1953 until year 1962 .}\\
    \end{tabular}
    \caption{An example sentence from the test set. Each translation has a corresponding gloss in italics. Errors are marked in bold.}
    \label{output-example}
  \end{center}
\end{figure*}

\subsection{Additional Language Pairs}

We experiment with translation from English into German, Polish, and Romanian.

Our English-German system is trained on the data available for the WMT14
translation task: Europarl \cite{europarl} and the Common Crawl
corpus,\footurl{http://commoncrawl.org/} roughly 4.3 million sentence pairs
altogether. We tune the system on the WMT13 test set and we test on the WMT14
set. We use TreeTagger \cite{treetagger} to lemmatize and tag the German data.

English-Polish has not been included in WMT shared tasks so far, but was present
as a language pair for several IWSLT editions which concentrate on TED talk
translation. Full test sets are only available for 2010, 2011, and 2012. The
references for 2013 and 2014 were not made public. We use the development set
and test set from 2010 as development data for parameter tuning. The remaining
two test sets (2011, 2012) are our test data. We train on the concatenation of
Europarl and WIT\textsuperscript{3} \cite{cettoloEtAl:EAMT2012}, ca. 750
thousand sentence pairs. The Polish half has been tagged using WCRFT
\cite{series/sci/Radziszewski13} which produces full morphological tags
compatible with the NKJP tagset \cite{prze:09b}. 

English-Romanian was added in WMT16. We train our system using the available
parallel data -- Europarl and SETIMES2 \cite{opus-setimes}, roughly 600 thousand
sentence pairs. We tune the English-Romanian system on the official development
set and we test on the WMT16 test set. We use the online tagger by
\newcite{romanian-tagger} to pre-process the data.

\begin{table}[h]
  \begin{center}
    \begin{tabular}{l|cccc}
      language & de & pl (2011) & pl (2012) & ro \\
      \hline 
      baseline & 15.7 & 12.8 & 10.4 & 19.6 \\
      +target & \bf 16.2 & \bf 13.4 & \bf 11.1 & \bf 20.2 \\
    \end{tabular}
    \caption{BLEU scores of the baseline and of the full model for English to
    German, Polish, and Romanian.}
    \label{vwscores-otherlangs}
  \end{center}
\end{table}

\Tref{vwscores-otherlangs} shows the obtained results. Similarly to
English-Czech experiments, BLEU scores are averaged over 5 independent
optimization runs.
Our system outperforms the baseline by 0.5-0.7 BLEU points in all cases, showing
that the method is applicable to other languages with rich morphology.

%
%

\begin{figure*}[!t]
  \begin{center}
    \begin{small}
    \begin{tabular}{rl}
      \bf input: & destruction of the equipment means that Syria can no longer produce new chemical weapons .\\
      \bf +source: & \textbf{zničením} zařízení znamená , že Sýrie již nemůže vytvářet nové chemické zbraně . \\
                   & \emph{\textbf{destruction\_of$_{instr}$} equipment means , that Syria already cannot produce new chemical weapons .} \\
      \bf +target: & zničení zařízení znamená , že Sýrie již nemůže vytvářet nové chemické zbraně . \\
                   & \emph{destruction\_of$_{nom}$ equipment means , that Syria already cannot produce new chemical weapons .} \\
      \hline
      \bf input: & nothing like that existed , and despite that we knew far more about each other .\\
      \bf +source: & nic takového neexistovalo , a přesto jsme věděli daleko víc o \textbf{jeden na druhého} . \\
                   & \emph{nothing like\_that existed , and despite\_that we knew far more about \textbf{one$_{nom}$ on other}} .\\
      \bf +target: & nic takového neexistovalo , a přesto jsme věděli daleko víc o sobě navzájem .\\
                  & \emph{nothing like\_that existed , and despite\_that we knew far more about each other .} \\
      \hline
      \bf input: & the authors have been inspired by their neighbours .\\
      \bf +source: & autoři byli inspirováni \textbf{svých sousedů} .\\
                   & \emph{the authors have been inspired \textbf{their$_{gen}$ neighbours$_{gen}$} .} \\
      \bf +target: & autoři byli inspirováni svými sousedy .\\
                   & \emph{the authors have been inspired their$_{instr}$ neighbours$_{instr}$ .}\\
    \end{tabular}
  \end{small}
    \caption{Example sentences from the test set showing improvements in morphological coherence. Each translation has a corresponding gloss in italics. Errors are marked in bold.}
    \label{output-coherence}
  \end{center}
\end{figure*}

\section{Analysis}

We manually analyze the outputs of English-Czech systems.
\Fref{output-example} shows an example sentence from the WMT14 test set
translated by all the system variants. The baseline system makes an error in
verb valency; the Czech verb ``došlo'' could be used but this verb already has
an (implicit) subject and the translation of ``mining'' (``těžba'') would have to be in a different
case and at a different position in the sentence. The second error is more
interesting, however: the baseline system fails to correctly identify the word sense of the
particle ``to'' and translates it in the sense of purpose, as in ``in order
to''. The source-context model takes the context (span of years) into
consideration and correctly disambiguates the translation of ``to'', choosing
the temporal meaning. It still fails to translate the main verb correctly,
though. Only the full model with target-context information is able to also
correctly translate the verb and inflect its arguments according to their roles
in the valency frame. The translation produced by this final system in this case
is almost flawless.

In order to verify that the automatically measured results correspond to visible
improvements in translation quality, we carried out two annotation experiments.
We took a random sample of 104 sentences from the test set and blindly ranked
two competing translations (the selection of sentences was identical for both
experiments).  In the first experiment, we compared the baseline system with
+source. In the other experiment, we compared the baseline with +target.  The
instructions for annotation were simply to compare overall translation quality;
we did not ask the annotator to look for any specific phenomena.  In terms of
automatic measures, our selection has similar characteristics as the full test
set: BLEU scores obtained on our sample are 15.08, 16.22 and 16.53 for the
baseline, +source and +target respectively.

In the first case, the annotator marked 52 translations as equal in quality, 26
translations produced by +source were marked as better and in the remaining 26
cases, the baseline won the ranking. Even though there is a difference in BLEU,
human annotation does not confirm this measurement, ranking both systems
equally.

In the second experiment, 52 translations were again marked as equal. In 34
cases, +target produced a better translation while in 18 cases, the baseline
output won. The difference between the baseline and +target suggests that the
target-context model may provide information which is useful for translation
quality as perceived by humans.

Our overall impression from looking at the system outputs was that both the
source-context and target-context model tend to fix many morpho-syntactic
errors. Interestingly, we do not observe as many improvements in the word/phrase
sense disambiguation, though the source context does help semantics in
some sentences. The target-context model tends to preserve the overall
agreement and coherence better than the system with a source-context model only. We
list several such examples in \Fref{output-coherence}. Each of them is fully
corrected by the target-context model, producing an accurate translation of the
input.

%
%

\section{Related Work}

\label{related-work}

Discriminative models in MT have been proposed before. \newcite{carpuat:07b}
trained a maximum entropy classifier for each source phrase type which used
source context information to disambiguate its translations. The models did not
capture target-side information and they were independent; no parameters were
shared between classifiers for different phrases. They used a strong feature set
originally developed for word sense disambiguation.
\newcite{gimpel-08} also used wider source-context information but did not train
a classifier; instead, the features were included directly in the log-linear
model of the decoder.
\newcite{mauser2009} introduced the ``discriminative word lexicon'' and trained
a binary classifier for each target word, using as features only the bag of
words (from the whole source sentence). Training sentences where the target word
occurred were used as positive examples, other sentences served as negative
examples. 
\newcite{jeong2010} proposed a discriminative lexicon with a rich feature set
tailored to translation into morphologically rich languages; unlike our work,
their model only used source-context features.

\newcite{subotin11} included target-side context information in a
maximum-entropy model for the prediction of morphology. The work was done
within the paradigm of hierarchical PBMT and assumes that cube pruning is used
in decoding.  Their algorithm was tailored to the specific problem of passing
non-local information about morphological agreement required by individual
rules (such as explicit rules enforcing subject-verb agreement).  Our algorithm
only assumes that hypotheses are constructed left to right and provides a
general way for including target context information in the classifier,
regardless of the type of features.  Our implementation is freely available and
can be further extended by other researchers in the future.

%
%

\section{Conclusions}

We presented a discriminative model for MT which uses both source and target
context information. We have shown that such a model can be used directly during
decoding in a relatively efficient way. We have shown that this model
consistently significantly improves the quality of English-Czech translation
over a strong baseline with large training data. We have validated the
effectiveness of our model on several additional language pairs. We have
provided an analysis showing concrete examples of improved lexical selection and
morphological coherence. Our work is available in the main branch of Moses for
use by other researchers.

\section*{Acknowledgements}

This work has received funding from the European Union's Horizon 2020 research
and innovation programme under grant agreements no. 644402 (HimL) and 645452
(QT21), from the European Research Council (ERC) under grant agreement no.
640550, and from the SVV project number 260 333. This work has been using
language resources stored and distributed by the LINDAT/CLARIN project of the
Ministry of Education, Youth and Sports of the Czech Republic (project
LM2015071). 

\bibliographystyle{acl2016}
\bibliography{biblio}

\end{document}